\documentclass[sigconf]{acmart}
\usepackage{bbm}
\usepackage{multirow}
\usepackage{graphicx}
\usepackage{hhline}
\newcommand*\rot{\rotatebox{90}}
\usepackage{subcaption}
\newcommand{\scheme}{TOG}
\renewcommand\footnotetextcopyrightpermission[1]{}

\AtBeginDocument{%
	\providecommand\BibTeX{{%
			\normalfont B\kern-0.5em{\scshape i\kern-0.25em b}\kern-0.8em\TeX}}}

\setcopyright{none}
\acmConference[]{Technical Report}{February 28, 2020}{Georgia Institute of Technology}

\makeatletter
\newcommand*{\centerfloat}{%
	\parindent \z@
	\leftskip \z@ \@plus 1fil \@minus \textwidth
	\rightskip\leftskip
	\parfillskip \z@skip}
\makeatother

\begin{document}
	
	\title[\scheme{}: Targeted Adversarial Objectness Gradient Attacks on Real-time Object Detection Systems]{\scheme{}: Targeted Adversarial Objectness Gradient Attacks \\ on Real-time Object Detection Systems}
	\titlenote{This work is released as a technical report at the Distributed Data Intensive Systems Lab (DiSL), Georgia Institute of Technology, with the following version history\\
		\textbf{v1}: November 15, 2019\\
		\textbf{v2}: February 28, 2020\\
	Readers may visit \url{https://github.com/git-disl/TOG} for updates.}
	
	\author{Ka-Ho Chow, Ling Liu, Mehmet Emre Gursoy, Stacey Truex, Wenqi Wei, Yanzhao Wu}
	\affiliation{%
		\institution{Georgia Institute of Technology}
		\city{Atlanta}
		\state{Georgia}
		\postcode{30332}
	}
	
	\begin{abstract}
		The rapid growth of real-time huge data capturing has pushed the deep learning and data analytic computing to the edge systems. Real-time object recognition on the edge is one of the representative deep neural network (DNN) powered edge systems for real-world mission-critical applications, such as autonomous driving and augmented reality. While DNN powered object detection edge systems celebrate many life-enriching opportunities, they also open doors for misuse and abuse. This paper presents three Targeted adversarial Objectness Gradient attacks, coined as \scheme{}, which can cause the state-of-the-art deep object detection networks to suffer from object-vanishing, object-fabrication, and object-mislabeling attacks. We also present a universal objectness gradient attack to use adversarial transferability for black-box attacks, which is effective on any inputs with negligible attack time cost, low human perceptibility, and particularly detrimental to object detection edge systems. We report our experimental measurements using two benchmark datasets (PASCAL VOC and MS COCO) on two state-of-the-art detection algorithms (YOLO and SSD). The results demonstrate serious adversarial vulnerabilities and the compelling need for developing robust object detection systems.
	\end{abstract}
	
	
	\maketitle
	
	\section{Introduction}
\begin{figure}
		\centering
		\includegraphics[width=0.95\columnwidth]{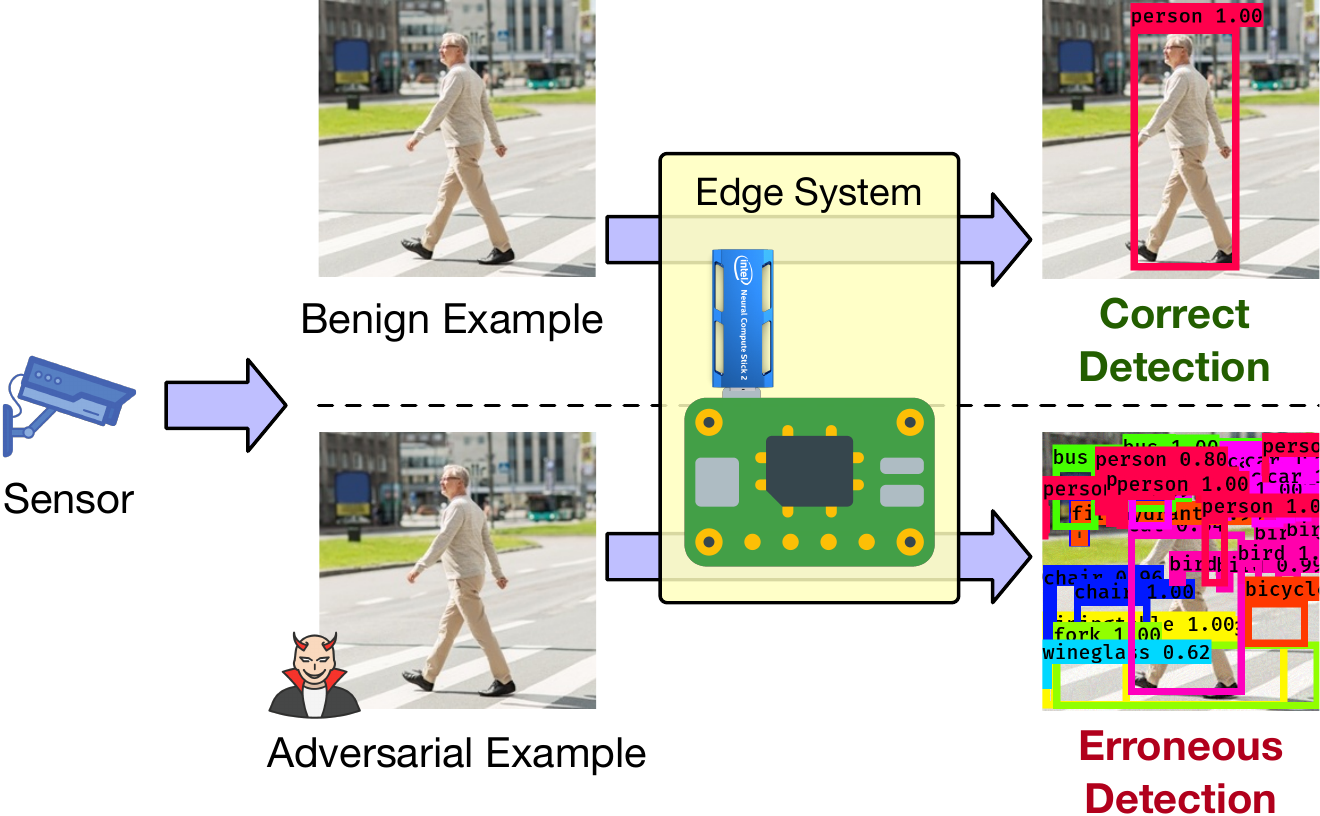}
		\vspace{-0.5em}
		\caption{The edge system correctly identifies the person on the benign input (top) but misdetects given the adversarial example (bottom), which is visually indistinguishable from the benign one.}
		\label{fig:edge-overview}
		\vspace{-0.5em}
	\end{figure}

	Edge data analytics and deep learning as a service on the edge have attracted a flurry of research and development efforts in both industry and academics~\cite{reagen2016minerva,ncs2}. Open source deep object detection networks~\cite{ren2015faster,liu2016ssd,redmon2018yolov3} have fueled new edge applications and edge system deployments, such as traffic sign identification on autonomous vehicles~\cite{simon2019complexer} and intrusion detection on smart surveillance systems~\cite{gajjar2017human}. However, very few performed systematic studies on the vulnerabilities of real-time deep object detectors, which are critical to edge security and privacy. Figure~\ref{fig:edge-overview} shows a typical scenario where an edge system receives an input image or video frame from a sensor (e.g., a camera), and it runs a real-time DNN object detection model (e.g., YOLOv3~\cite{redmon2018yolov3}) on the edge device (e.g., a Raspberry Pi with an AI acceleration module). With no attack, the well-trained object detector can process the benign input (top) and accurately identify a person walking across the street. However, under attack with an adversarial example (bottom), which is visually indistinguishable by human perception to the benign input, the \emph{same} object detector will be fooled to make erroneous detection.
	
\begin{figure}
		\centering
		\includegraphics[width=0.93\columnwidth]{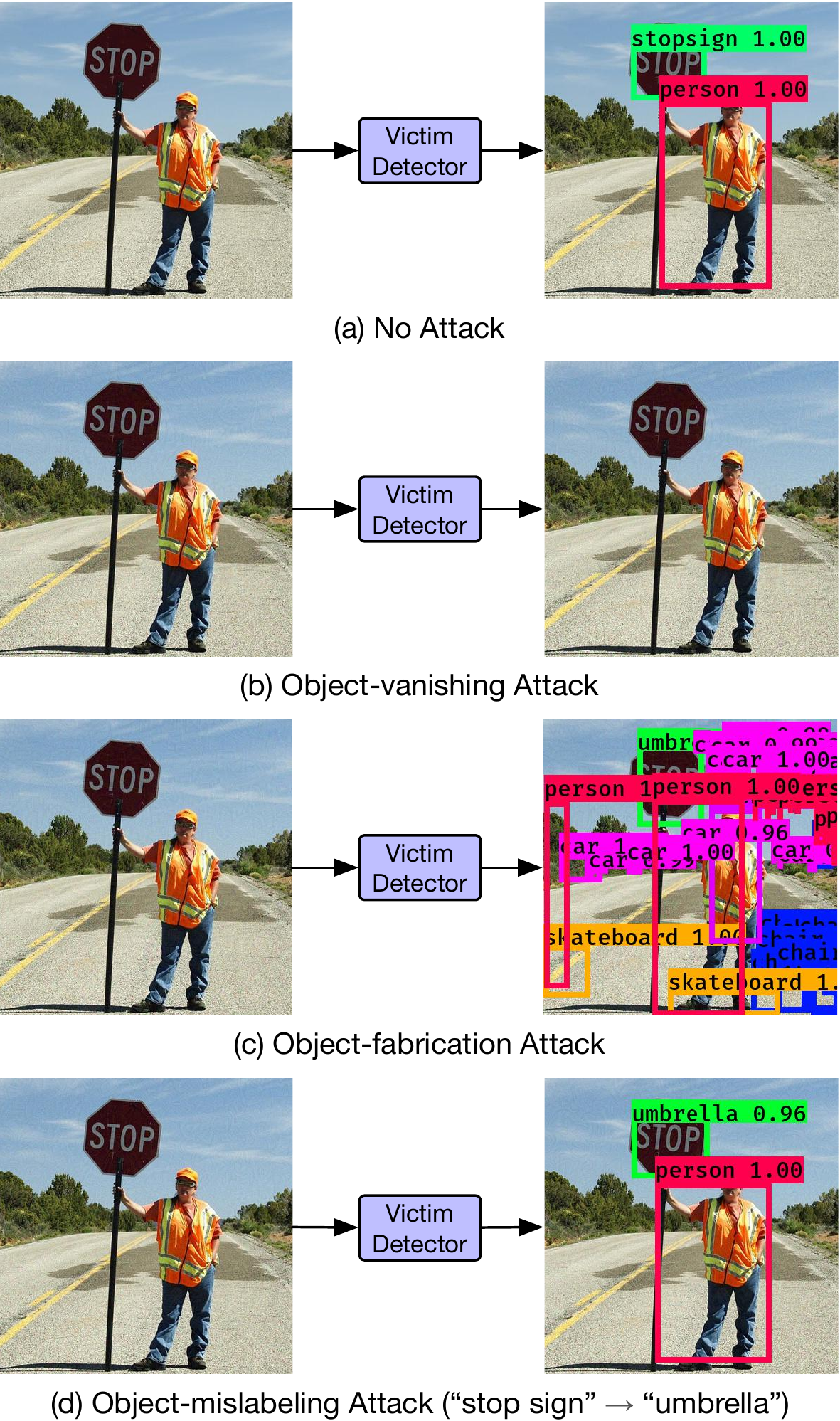}	
		\vspace{-0.5em}	
		\caption{Illustration of three \scheme{} attacks (2nd-4th rows) to deep object detection networks. \emph{Left}: Benign input. \emph{Right}: Detection results under the adversarial attacks.}
		\label{fig:showcase}
	\end{figure}

	In this paper, we present three vulnerabilities of DNN object detection systems, by developing three \textbf{T}argeted adversarial \textbf{O}bjectness \textbf{G}radient attacks, as a family of \scheme{} attacks on real-time object detection systems. Although there is a large body of adversarial attacks to DNN image classifiers~\cite{goodfellow2014explaining} in literature, they are mainly effective in causing DNN classifiers to produce wrong classifications by using different attack strategies to determine the location and the amount of per-pixel perturbation to inject to a benign input image~\cite{wei2018adversarial}. In contrast, deep object detection networks detect and segment multiple objects that may be visually overlapping in a single image or video frame and provide one class label to each of the detected objects. Thus, an in-depth understanding of various vulnerabilities of deep object detectors is more complicated than misclassification attacks in the DNN image classifiers, because the DNN object detectors have larger and varying attack surfaces, such as the object existence, object location, and object class label, which open more opportunities for attacks with various adversarial goals and sophistications. The \scheme{} attacks are the first targeted adversarial attack method on object detection networks by targeting at different objectness semantics, such as making objects vanishing, fabricating more objects, or mislabeling some or all objects. Each of these attacks injects a human-imperceptible adversarial perturbation to fool a real-time object detector to misbehave in three different ways, as shown in Figure~\ref{fig:showcase}. The \emph{object-vanishing} attack in Figure~\ref{fig:showcase}(b) causes all objects to vanish from the YOLOv3~\cite{redmon2018yolov3} detector. The \emph{object-fabrication} attack in Figure~\ref{fig:showcase}(c) causes the detector to output many false objects with high confidence. The \emph{object-mislabeling} attack in Figure~\ref{fig:showcase}(d) fools the detector to mislabel (e.g., the stop sign becomes an umbrella). 
	We further present a highly efficient universal adversarial perturbation algorithm that generates a single universal perturbation that can perturb any input to fool the victim detector effectively. Given the attack is generated offline, by utilizing adversarial transferability, one can use the \scheme{} universal perturbation to launch a black-box attack with a very low (almost zero) online attack cost, which is particularly fatal in real-time edge applications for object detections. 
	
	The rest of the paper is organized as follows. We first present the \scheme{} attacks in Section~\ref{sec:methodology}, then experimental evaluations in Section~\ref{sec:experiment}, followed by concluding remarks in Section~\ref{sec:conclusion}.
	
	\section{\scheme{} Attacks}\label{sec:methodology}  
	Deep object detection networks, in general, share the same input-output structure with a similar formulation. They all take the input image or video frame and produce outputs by using bounding box techniques to provide object localization for all objects of interest and by giving the classification for each detected object~\cite{redmon2016you,redmon2017yolo9000,redmon2018yolov3,ren2015faster,liu2016ssd}. The \scheme{} attacks are constructed without restriction to any particular detection algorithm, as shown in our experimental evaluation. To illustrate the details of \scheme{} attacks, we choose to use YOLOv3~\cite{redmon2018yolov3} to present our formulation in this section.
	
	Given an input image $\boldsymbol{x}$, the object detector first detects a large number of $S$ candidate bounding boxes $\hat{\boldsymbol{\mathcal{B}}}(\boldsymbol{x})=\{\hat{\boldsymbol{o}}_1,\hat{\boldsymbol{o}}_2,\dots,\hat{\boldsymbol{o}}_S\}$ where $\hat{\boldsymbol{o}}_i=(\hat{b}^x_i, \hat{b}^y_i, \hat{b}^W_i, \hat{b}^H_i, \hat{C}_i, \hat{\boldsymbol{p}}_i)$ is a candidate centered at $(\hat{b}^x_i, \hat{b}^y_i)$ having a dimension $(\hat{b}^W_i, \hat{b}^H_i)$ with a probability of $\hat{C}_i\in[0,1]$ having an object contained, and a $K$-class probabilities $\hat{\boldsymbol{p}}_i=(\hat{p}^1_i, \hat{p}^2_i,\dots,\hat{p}^K_i)$. This is done by dividing the input $\boldsymbol{x}$ into mesh grids in different scales (resolutions) where each grid cell produces multiple candidate bounding boxes based on the anchors and is responsible for locating objects centered at the cell. The final detection results $\hat{\boldsymbol{\mathcal{O}}}(\boldsymbol{x})$ are obtained by applying confidence thresholding to remove candidates with low prediction confidence (i.e., $\max_{1\leq c\leq K}\hat{C}_i\hat{p}^c_i$) and non-maximum suppression to exclude those with high overlapping.
	
	An adversarial example $\boldsymbol{x}'$ is generated by perturbing a benign input $\boldsymbol{x}$ sent to the victim detector, aiming to fool the victim to misdetect randomly (untargeted) or purposefully (targeted). The generation process of the adversarial example can be formulated as
	\begin{equation}\label{eq:adv-process}
	\min||\boldsymbol{x}' - \boldsymbol{x}||_{p} \quad s.t.\;\hat{\boldsymbol{\mathcal{O}}}(\boldsymbol{x}')=\boldsymbol{\mathcal{O}}^*, \hat{\boldsymbol{\mathcal{O}}}(\boldsymbol{x}')\neq\hat{\boldsymbol{\mathcal{O}}}(\boldsymbol{x}),
	\end{equation}
	where $p$ is the distance metric, which can be the $L_0$ norm measuring the percentage of the pixels that are changed, the $L_2$ norm computing the Euclidean distance, or the $L_\infty$ norm denoting the maximum change to any pixel, and $\boldsymbol{\mathcal{O}}^*$ denotes the target detections for targeted attacks or any incorrect ones for untargeted attacks. 
	
\begin{figure*}
		\centering
		\includegraphics[width=0.90\linewidth]{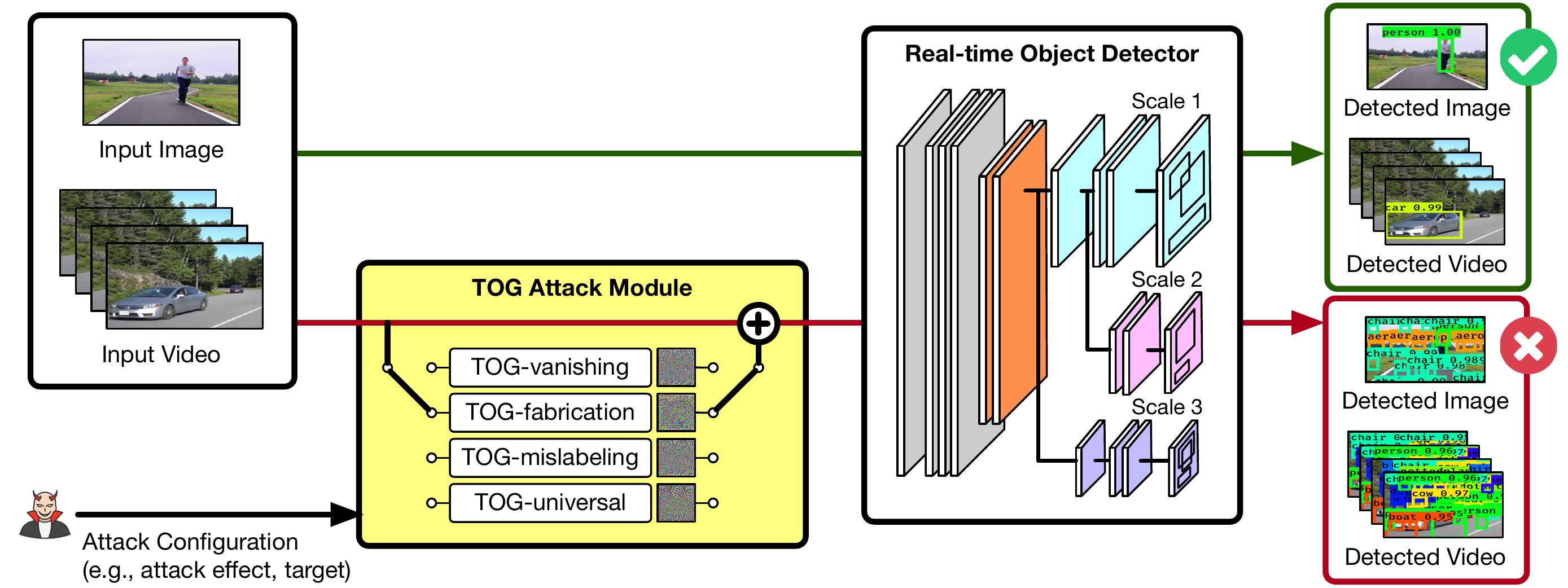}
		\vspace{-0.5em}	
		\caption{{An illustration of the adversarial attacks using \scheme{}.}}
		\label{fig:system}
		\vspace{-0.5em}
	\end{figure*}

	Figure~\ref{fig:system} illustrates the adversarial attacks using \scheme{}. 
	Given an input source (e.g., an image or video frame), \scheme{} attack module takes the configuration specified by the adversary to prepare for the corresponding adversarial perturbation, which will be added to the input to cause the victim to misdetect. The first three attacks in \scheme{}: \scheme{}-vanishing, \scheme{}-fabrication, and \scheme{}-mislabeling tailor an adversarial perturbation for each input, while \scheme{}-universal uses the same universal perturbation to corrupt any input. 
	
	Training a deep neural network often starts with random initialization of model weights, which will be updated slowly by taking the derivative of the loss function $\mathcal{L}$ with respect to the learnable model weights $\boldsymbol{\mathcal{W}}$ over a mini-batch of input-output pairs $\{(\tilde{\boldsymbol{x}}, \boldsymbol{\mathcal{O}})\}$ with the following equation until convergence:
	\begin{equation}
	\boldsymbol{\mathcal{W}}_{t+1}=\boldsymbol{\mathcal{W}}_t - \alpha\frac{\partial\; \mathbb{E}_{(\tilde{\boldsymbol{x}}, \boldsymbol{\mathcal{O}})}[\mathcal{L}(\tilde{\boldsymbol{x}};\boldsymbol{\mathcal{O}}, \boldsymbol{\mathcal{W}}_t)]}{\partial \boldsymbol{\mathcal{W}}_t}
	\end{equation}
	where $\alpha$ is the learning rate controlling the step size of the update. While training deep object detection networks is done by \emph{fixing} the input image $\tilde{\boldsymbol{x}}$ and progressively \emph{updating} the model weights $\boldsymbol{\mathcal{W}}$ towards the goal defined by the loss function, \scheme{} conducts adversarial attacks by reversing the training process. We \emph{fix} the model weights of the victim detector and iteratively \emph{update} the input image $\boldsymbol{x}$ towards the goal defined by the type of the attack to be launched with the following general equation:
	\begin{equation}\label{eq:attack}
	\boldsymbol{x}'_{t+1}=\prod\limits_{\boldsymbol{x}, \epsilon}\bigg[\boldsymbol{x}'_t - \alpha\Gamma\bigg(\frac{\partial \mathcal{L}^*(\boldsymbol{x}'_t;\boldsymbol{\mathcal{O}}^*, \boldsymbol{\mathcal{W}})}{\partial\boldsymbol{x}'_t}\bigg)\bigg]
	\end{equation}
	where $\prod_{\boldsymbol{x},\epsilon}[\cdot]$ is the projection onto a hypersphere with a radius $\epsilon$ centered at $\boldsymbol{x}$ in $L_p$ norm, $\Gamma$ is a sign function, and $\mathcal{L}^*$ defines the loss function to be optimized during the attack.
	
	In deep object detection networks, every ground-truth object in a training sample $\tilde{\boldsymbol{x}}$ will be assigned to one of the $S$ bounding boxes according to the center coordinates and the amount of overlapping with the anchors. Let $\mathbbm{1}_i=1$ if the $i$-th bounding box is responsible for an object and $0$ otherwise. Then, $\boldsymbol{\mathcal{O}}=\{\boldsymbol{o}_i\vert \mathbbm{1}_i=1, 1\leq i\leq S\}$ is a set of ground-truth objects where $\boldsymbol{o}_i=(b^x_i, b^y_i, b^W_i, b^H_i, \boldsymbol{p}_i)$ with $\boldsymbol{p}_i=(p^1_i,p^2_i,\dots,p^K_i)$ and $p^c_i=1$ if $\boldsymbol{o}_i$ is a class $c$ object. The optimization objective of the deep object detection network consists of three parts, each of them corresponds to one of the three output structures describing a detected object (i.e., existence, locality, and class label). The objectness score $\hat{C}_i\in[0,1]$ determines the existence of an object in the candidate bounding box, which can be learned by minimizing the binary cross-entropy $\ell_{\text{BCE}}$:
	\begin{equation}
	\begin{split}
	\mathcal{L}_{\text{obj}}(\tilde{\boldsymbol{x}};\boldsymbol{\mathcal{O}}, \boldsymbol{\mathcal{W}})&=\sum_{i=1}^S\bigg[\mathbbm{1}_i\ell_{\text{BCE}}(1, \hat{C}_i)\bigg] \\
	\mathcal{L}_{\text{noobj}}(\tilde{\boldsymbol{x}};\boldsymbol{\mathcal{O}}, \boldsymbol{\mathcal{W}})&=\sum_{i=1}^S\bigg[(1-\mathbbm{1}_i)\ell_{\text{BCE}}(0, \hat{C}_i)\bigg]
	\end{split}
	\end{equation}
	The center coordinates $(\hat{b}^x_i, \hat{b}^y_i)$ and dimension $(\hat{b}^W_i, \hat{b}^H_i)$ give the spatial locality, learned by minimizing the squared error $\ell_{\text{SE}}$:
	\begin{equation}
	\begin{split}
	\mathcal{L}_{\text{loc}}(\tilde{\boldsymbol{x}};\boldsymbol{\mathcal{O}, \boldsymbol{\mathcal{W}}})
	= & \sum_{i=1}^S\mathbbm{1}_i\big[\ell_{\text{SE}}\big(b^x_i,\hat{b}^x_i\big)+\ell_{\text{SE}}\big(b^y_i,\hat{b}^y_i\big) \\
	+ & \ell_{\text{SE}}\big(\sqrt{b^W_i},\sqrt{\hat{b}^W_i}\big) + \ell_{\text{SE}}\big(\sqrt{b^H_i},\sqrt{\hat{b}^H_i}\big)\big]
	\end{split}
	\end{equation} 
	The last part is the $K$-class probabilities $\hat{\boldsymbol{p}}_i=(\hat{p}^1_i, \hat{p}^2_i,\dots,\hat{p}^K_i)$ that estimate the class label of the corresponding candidate, optimized by minimizing the binary cross-entropy:
	\begin{equation}
	\mathcal{L}_{\text{prob}}(\tilde{\boldsymbol{x}};\boldsymbol{\mathcal{O}}, \boldsymbol{\mathcal{W}})=\sum_{i=1}^S\mathbbm{1}_i\sum_{c\in\text{classes}}\ell_{\text{BCE}}\big(p^c_i,\hat{p}^c_i\big)
	\end{equation}
	As a result, the deep object detection network can be optimized by the linear combination of the above loss functions:
	\begin{equation}\label{eq:loss}
	\begin{split}
	\mathcal{L}(\tilde{\boldsymbol{x}};\boldsymbol{\mathcal{O}}, \boldsymbol{\mathcal{W}}) = & \mathcal{L}_{\text{obj}}(\tilde{\boldsymbol{x}};\boldsymbol{\mathcal{O}}, \boldsymbol{\mathcal{W}}) + \lambda_{\text{noobj}}\mathcal{L}_{\text{noobj}}(\tilde{\boldsymbol{x}};\boldsymbol{\mathcal{O}}, \boldsymbol{\mathcal{W}}) \\
	&+ \lambda_{\text{loc}}\mathcal{L}_{\text{loc}}(\tilde{\boldsymbol{x}};\boldsymbol{\mathcal{O}}, \boldsymbol{\mathcal{W}}) + \mathcal{L}_{\text{prob}}(\tilde{\boldsymbol{x}};\boldsymbol{\mathcal{O}}, \boldsymbol{\mathcal{W}})
	\end{split}
	\end{equation}
	where $\lambda_{\text{noobj}}$ and $\lambda_{\text{loc}}$ are hyperparameters penalizing incorrect objectness scores and bounding boxes  respectively.
	\begin{table*}\small
		\centering
		\begin{tabular}{@{}clcccccc@{}}
			\toprule
			\multirow{2}{*}{\textbf{Dataset}} & \multicolumn{1}{c}{\multirow{2}{*}{\textbf{Detector}}} & \multirow{2}{*}{\textbf{\begin{tabular}[c]{@{}c@{}}Benign\\ mAP (\%)\end{tabular}}} & \multicolumn{5}{c}{\textbf{Adversarial mAP (\%)}} \\ \cmidrule(l){4-8} 
			& \multicolumn{1}{c}{} &  & \textbf{\scheme{}-vanishing} & \textbf{\scheme{}-fabrication} & \textbf{\scheme{}-mislabeling (ML)} & \textbf{\scheme{}-mislabeling (LL)} & \textbf{\scheme{}-universal} \\ \cmidrule{1-8}
			\multirow{4}{*}{VOC} & YOLOv3-D & 83.43 & 0.31 & 1.41 & 4.58 & 1.73 & 14.16 \\
			& YOLOv3-M & 72.51 & 0.58 & 3.37 & 3.75 & 0.45 & 20.44 \\
			& SSD300 & 78.09 & 5.58 & 7.34 & 2.81 & 0.78 & 31.80 \\
			& SSD512 & 79.83 & 9.90 & 8.34 & 2.22 & 0.77 & 46.17 \\ \cmidrule{1-8}
			COCO & YOLOv3-D & 54.89 & 0.51 & 5.89 & 7.02 & 0.90 & 12.73 \\ \bottomrule
		\end{tabular}
	\vspace{1em}
		\caption{ Evaluation of \scheme{} attacks to four victim detectors.}
		\label{tab:overall_map}
		\vspace{-1em}
	\end{table*}
	\begin{figure*}
		\centering
		\begin{subfigure}[t]{0.25\textwidth}
			\centering
			\includegraphics[width=\textwidth]{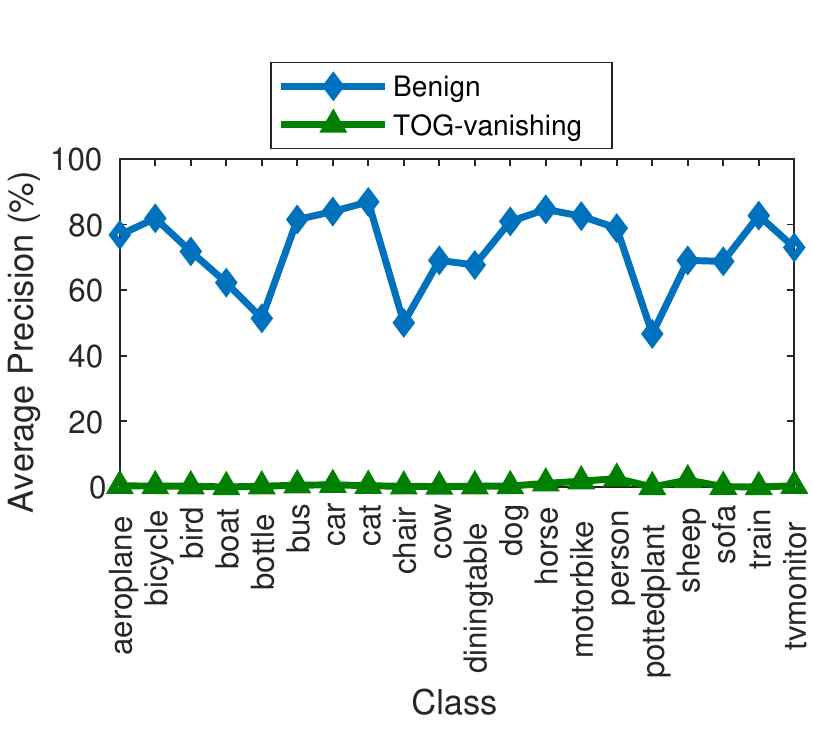}
			\caption{ \scheme{}-vanishing}
		\end{subfigure}%
		~ 
		\begin{subfigure}[t]{0.25\textwidth}
			\centering
			\includegraphics[width=\textwidth]{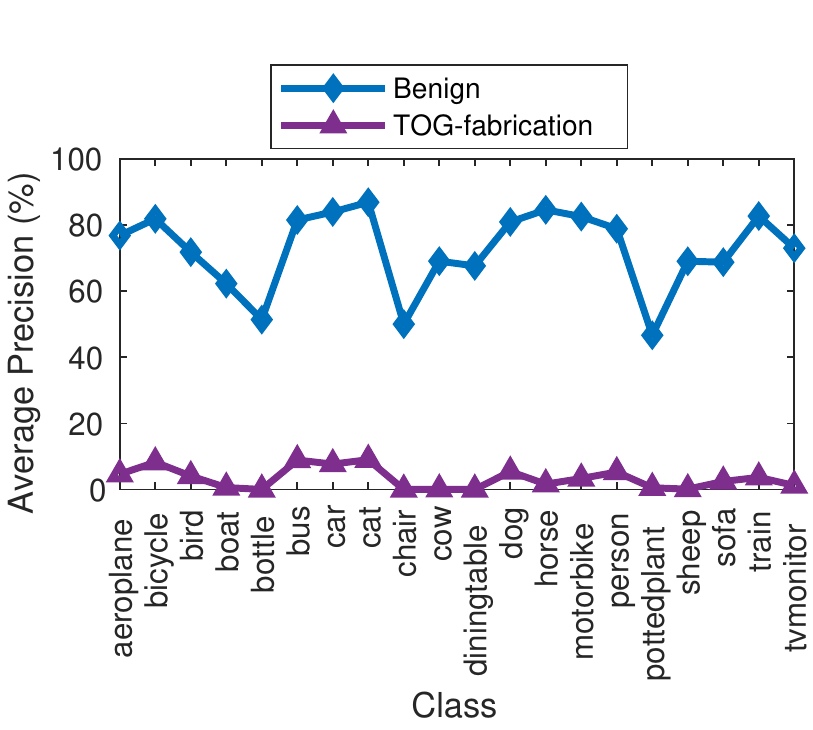}
			\caption{ \scheme{}-fabrication}
		\end{subfigure}%
		~ 
		\begin{subfigure}[t]{0.25\textwidth} 
			\centering
			\includegraphics[width=\textwidth]{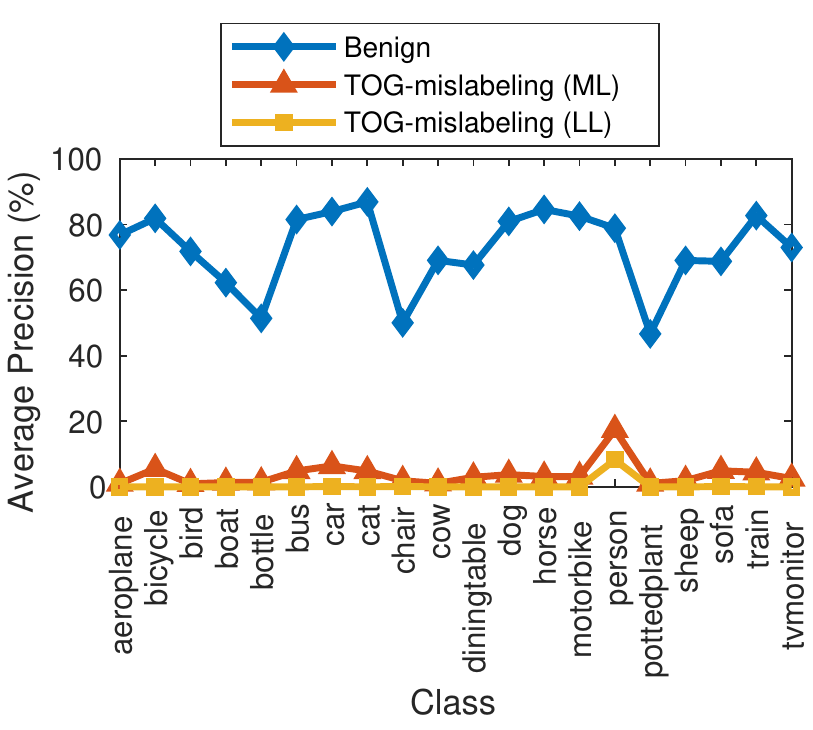}
			\caption{ \scheme{}-mislabeling}
		\end{subfigure}%
		~ 
		\begin{subfigure}[t]{0.25\textwidth}
			\centering
			\includegraphics[width=\textwidth]{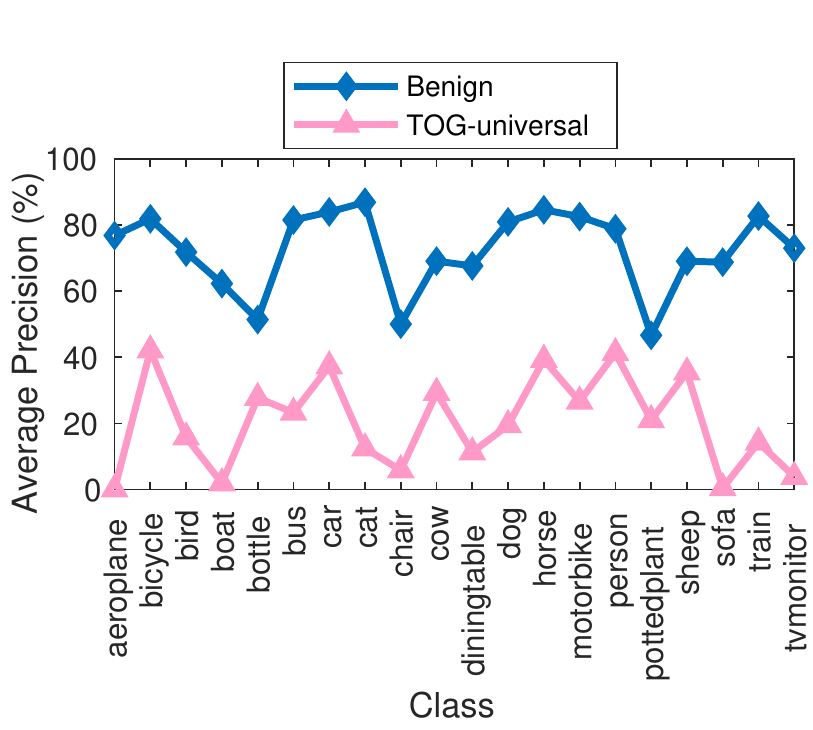}
			\caption{ \scheme{}-universal}
		\end{subfigure}%
		\vspace{-0.5em}
		\caption{ The average precision of each class in VOC for benign (blue) and adversarial examples generated by \scheme{} attacks.}
		\label{fig:exp-voc-yolov3-m}
	\end{figure*}
	
	To tailor an adversarial perturbation for each input $\boldsymbol{x}$ to generate the corresponding adversarial example $\boldsymbol{x}'$, \scheme{} is initialized with the benign example (i.e., $\boldsymbol{x}'_0=\boldsymbol{x}$) and sends the adversarial example $\boldsymbol{x}'_t$ at the $t$-th iteration to the victim detector to observe the detection results $\hat{\boldsymbol{\mathcal{O}}}(\boldsymbol{x}'_t)$. If the termination condition defined by the attack goal is achieved or the maximum number of iterations $T$ is reached, $\boldsymbol{x}'_t$ will be returned. Otherwise, it will be perturbed using Equation~\ref{eq:attack} to become $\boldsymbol{x}'_{t+1}$ for the new iteration.
	
	{\bf \scheme{}-vanishing.\/} For the \scheme{} object-vanishing attack, we set the target detection $\boldsymbol{\mathcal{O}}^*=\emptyset$ and $\mathcal{L}^*=\mathcal{L}$ to cause the victim to detect no objects on the adversarial example. 
	
	{\bf \scheme{}-fabrication.\/} For the \scheme{} object-fabrication attack, we set the target detection $\boldsymbol{\mathcal{O}}^*=\hat{\boldsymbol{\mathcal{O}}}(\boldsymbol{x})$ and $\mathcal{L}^*=-\mathcal{L}$ to return a large number of false objects. 
	
	{\bf \scheme{}-mislabeling.\/}  For the \scheme{} object-mislabeling attack, we set the target detection $\boldsymbol{\mathcal{O}}^*$ to be $\hat{\boldsymbol{\mathcal{O}}}(\boldsymbol{x})$ with each object having an incorrect label and $\mathcal{L}^*=\mathcal{L}$. While any incorrect class label can be assigned, we adopt a systematic approach to generate targets~\cite{wei2018adversarial}: the least-likely (LL) class attack picks the class label with the lowest probability (i.e., $y^*_i=\arg\min_{c}\hat{p}^c_i$), and the most-likely (ML) class attack finds the class label with the second-highest probability (i.e., $y^*_i=\arg\max_{c, \hat{p}^c_i\neq\max_u \hat{p}^u_i}\hat{p}^c_i$).
	
	{\bf \scheme{}-universal.\/} For the \scheme{} universal attack, the algorithm will generate a single universal perturbation for making any input suffer from an object-vanishing attack. Although all three \scheme{} attacks of generating adversarial perturbation for each input to fool an object detector are highly effective, they all require iterative optimization to produce effective perturbations during the online detection. The \scheme{} universal attack can generate a universal perturbation applicable to any input to the detector through iterative optimization by training. Let $\boldsymbol{\mathcal{D}}$ denote the set of $N$ training images, we want to gradually build the perturbation vector $\boldsymbol{\eta}$. At each iteration $t$, we obtain a training sample $\tilde{\boldsymbol{x}}\in\boldsymbol{\mathcal{D}}$ and find the additional perturbation $\Delta\boldsymbol{\eta}_t$ that causes the victim detector to make errors towards object vanishing attack goal in the current perturbed image $\tilde{\boldsymbol{x}}+\boldsymbol{\eta}_t$. We then add this additional perturbation $\Delta\boldsymbol{\eta}_t$ to the current universal adversarial perturbation $\boldsymbol{\eta}_t$ and clip the new perturbation to ensure the distortion is constrained within $[-\epsilon,\epsilon]$. The termination condition can be $\kappa\%$ of the objects in the training images vanish, or a maximum number of epochs $Q$ is reached. Upon completing the attack training, the universal perturbation can be applied to any given input to the real-time object detector running in an edge system. It is a black-box attack by adversarial transferability since it can be employed directly upon receiving a benign input to the object detector at runtime by only applying the pretrained perturbation. This attack can fool the object detector at an almost zero online attack time cost.
	
	\section{Experimental Evaluation}\label{sec:experiment}
	Extensive experiments are conducted using two popular benchmark datasets: PASCAL VOC~\cite{everingham2015pascal} and MS COCO~\cite{lin2014microsoft} on four state-of-the-art object detectors from two popular families of detection algorithms: ``YOLOv3-D" and ``YOLOv3-M" are the YOLOv3~\cite{redmon2018yolov3} models with a Darknet53 backbone and a MobileNetV1 backbone respectively. ``SSD300" and ``SSD512" are the SSD~\cite{liu2016ssd} models with an input resolution of $(300\times300)$ and $(512\times512)$ respectively. The VOC 2007+2012 dataset has $16,551$ training images and $4,952$ testing images. The COCO 2014 dataset has $117,264$ training images and $5,000$ testing images. We report the results on the entire test set. The mean average precision (mAP) of each dataset and victim detector is presented in the $3$rd column of Table~\ref{tab:overall_map}. All experiments use the default configurations from each detector without any fine-tuning of the hyperparameters in each setting. We produce adversarial perturbations in $L_\infty$ norm with a maximum distortion $\epsilon=0.031$, a step size $\alpha=0.008$, and the number of iterations $T=10$. For universal attacks, $12,800$ images from the training set are extract to form $\boldsymbol{\mathcal{D}}$ with a maximum distortion $\epsilon=0.031$, a learning rate $\alpha=0.0001$, and the number of training epochs $Q=50$. All attacks were conducted on NVIDIA RTX 2080 SUPER (8 GB) GPU with Intel i7-9700K (3.60GHz) CPU and 32 GB RAM on Ubuntu 18.04. 
	\begin{table*}\small
		\centerfloat
		\begin{tabular}{ccccccccc}
			\cline{1-1} \cline{3-9}
			\multicolumn{1}{|c|}{\multirow{3}{*}{\textbf{Input Image}}} & \multicolumn{1}{c|}{} & \multicolumn{1}{c|}{\multirow{3}{*}{}} & \multicolumn{6}{c|}{\textbf{Detection Results under four \scheme{} attacks}} \\ \cline{4-9} 
			\multicolumn{1}{|c|}{} & \multicolumn{1}{c|}{} & \multicolumn{1}{c|}{} & \multicolumn{1}{c|}{\multirow{2}{*}{\textbf{\begin{tabular}[c]{@{}c@{}}Benign\\ (No Attack)\end{tabular}}}} & \multicolumn{1}{c|}{\multirow{2}{*}{\textbf{\scheme{}-vanishing}}} & \multicolumn{1}{c|}{\multirow{2}{*}{\textbf{\scheme{}-fabrication}}} & \multicolumn{2}{c|}{\textbf{\scheme{}-mislabeling}} & \multicolumn{1}{c|}{\multirow{2}{*}{\textbf{\scheme{}-universal}}} \\ \cline{7-8} 
			\multicolumn{1}{|c|}{} & \multicolumn{1}{l|}{} & \multicolumn{1}{c|}{} & \multicolumn{1}{c|}{} & \multicolumn{1}{c|}{} & \multicolumn{1}{c|}{} & \multicolumn{1}{c|}{\textbf{ML}} & \multicolumn{1}{c|}{\textbf{LL}} & \multicolumn{1}{c|}{} \\ \cline{1-1} \cline{3-9} 
			\multicolumn{1}{|c|}{\includegraphics[scale=0.210]{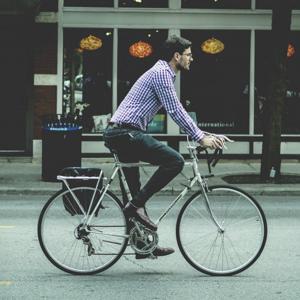}} & \multicolumn{1}{c|}{} & \multicolumn{1}{c|}{\rot{\;\;\;\;\;\;\;\;\textbf{SSD300}}} & \multicolumn{1}{c|}{\includegraphics[scale=0.210]{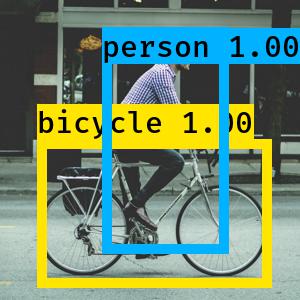}} & \multicolumn{1}{c|}{\includegraphics[scale=0.210]{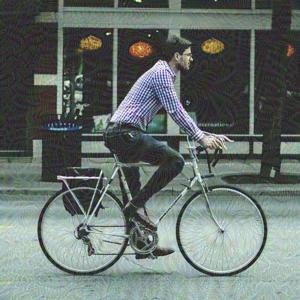}} & \multicolumn{1}{c|}{\includegraphics[scale=0.210]{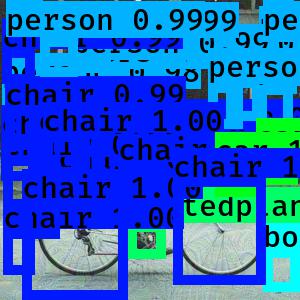}} & \multicolumn{1}{c|}{\includegraphics[scale=0.210]{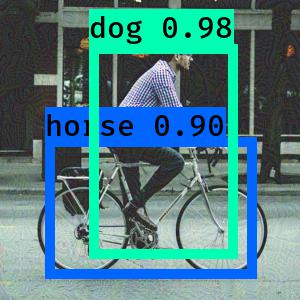}} & \multicolumn{1}{c|}{\includegraphics[scale=0.210]{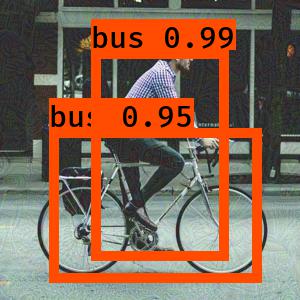}} & \multicolumn{1}{c|}{\includegraphics[scale=0.210]{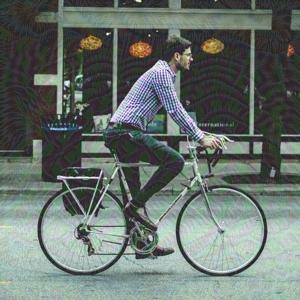}} \\ \cline{1-1} \cline{3-9}  \multicolumn{9}{c}{\textbf{=================================================================================================================}} \\ \cline{3-9} 
			\multicolumn{1}{c}{\multirow{3}{*}{}} & \multicolumn{1}{c|}{} & \multicolumn{1}{c|}{\multirow{3}{*}{}} & \multicolumn{6}{c|}{\textbf{Detection Results (Adversarial transferability of SSD300 to other victim detectors)}} \\ \cline{4-9} 
			\multicolumn{1}{c}{} & \multicolumn{1}{c|}{} & \multicolumn{1}{c|}{} & \multicolumn{1}{c|}{\multirow{2}{*}{\textbf{\begin{tabular}[c]{@{}c@{}}Benign\\ (No Attack)\end{tabular}}}} & \multicolumn{1}{c|}{\multirow{2}{*}{\textbf{\scheme{}-vanishing}}} & \multicolumn{1}{c|}{\multirow{2}{*}{\textbf{\scheme{}-fabrication}}} & \multicolumn{2}{c|}{\textbf{\scheme{}-mislabeling}} & \multicolumn{1}{c|}{\multirow{2}{*}{\textbf{\scheme{}-universal}}} \\ \cline{7-8} 
			\multicolumn{1}{c}{} & \multicolumn{1}{l|}{} & \multicolumn{1}{c|}{} & \multicolumn{1}{c|}{} & \multicolumn{1}{c|}{} & \multicolumn{1}{c|}{} & \multicolumn{1}{c|}{\textbf{ML}} & \multicolumn{1}{c|}{\textbf{LL}} & \multicolumn{1}{c|}{} \\ \cline{3-9} 
			& \multicolumn{1}{c|}{} & \multicolumn{1}{c|}{\rot{\textbf{\;\;\;\;\;SSD512}}} & \multicolumn{1}{c|}{\begin{tabular}[c]{@{}c@{}}\includegraphics[scale=0.210]{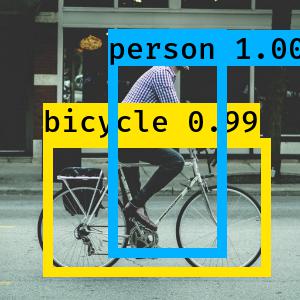}\\ \;\end{tabular}} & \multicolumn{1}{c|}{\begin{tabular}[c]{@{}c@{}}\includegraphics[scale=0.210]{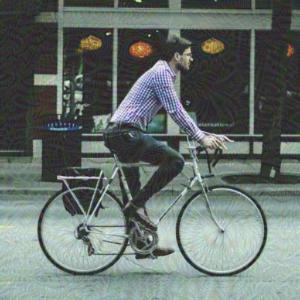}\\ \textbf{Transfer Succeeds}\end{tabular}} & \multicolumn{1}{c|}{\begin{tabular}[c]{@{}c@{}}\includegraphics[scale=0.210]{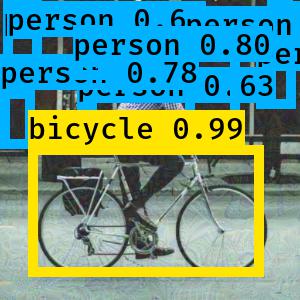}\\ \textbf{Transfer Succeeds}\end{tabular}} & \multicolumn{1}{c|}{\begin{tabular}[c]{@{}c@{}}\includegraphics[scale=0.210]{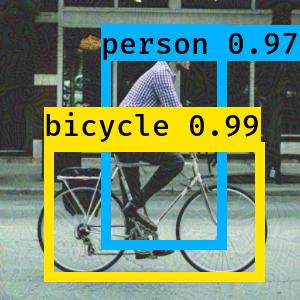}\\ \textbf{Transfer Fails}\end{tabular}} & \multicolumn{1}{c|}{\begin{tabular}[c]{@{}c@{}}\includegraphics[scale=0.210]{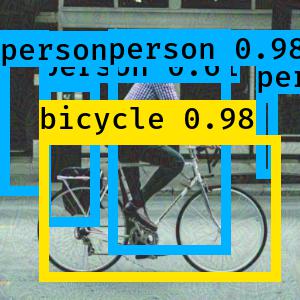}\\ \textbf{Transfer Succeeds}\end{tabular}} & \multicolumn{1}{c|}{\begin{tabular}[c]{@{}c@{}}\includegraphics[scale=0.210]{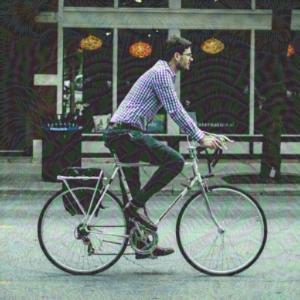}\\ \textbf{Transfer Succeeds}\end{tabular}} \\ \cline{3-9} 
			& \multicolumn{1}{c|}{} & \multicolumn{1}{c|}{\rot{\textbf{YOLOv3-D}}} & \multicolumn{1}{c|}{\begin{tabular}[c]{@{}c@{}}\includegraphics[scale=0.210]{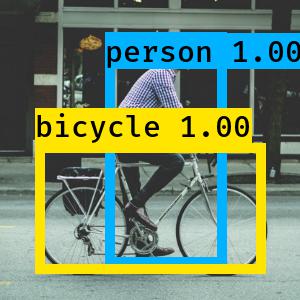}\\ \;\end{tabular}} & \multicolumn{1}{c|}{\begin{tabular}[c]{@{}c@{}}\includegraphics[scale=0.210]{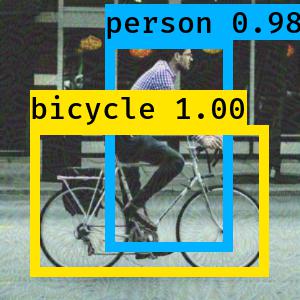}\\ \textbf{Transfer Fails}\end{tabular}} & \multicolumn{1}{c|}{\begin{tabular}[c]{@{}c@{}}\includegraphics[scale=0.210]{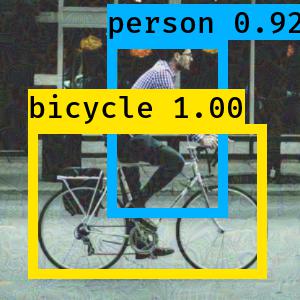}\\ \textbf{Transfer Fails}\end{tabular}} & \multicolumn{1}{c|}{\begin{tabular}[c]{@{}c@{}}\includegraphics[scale=0.210]{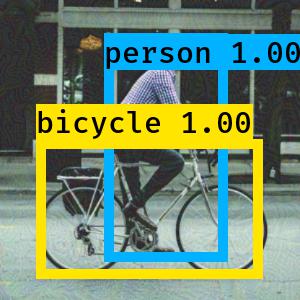}\\ \textbf{Transfer Fails}\end{tabular}} & \multicolumn{1}{c|}{\begin{tabular}[c]{@{}c@{}}\includegraphics[scale=0.210]{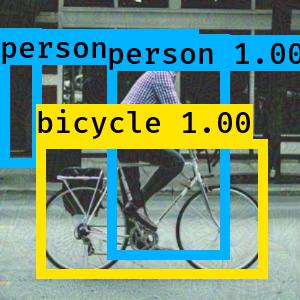}\\ \textbf{Transfer Succeeds}\end{tabular}} & \multicolumn{1}{c|}{\begin{tabular}[c]{@{}c@{}}\includegraphics[scale=0.210]{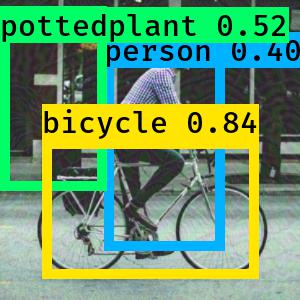}\\ \textbf{Transfer Succeeds}\end{tabular}} \\ \cline{3-9} 
			& \multicolumn{1}{c|}{} & \multicolumn{1}{c|}{\rot{\textbf{YOLOv3-M}}} & \multicolumn{1}{c|}{\begin{tabular}[c]{@{}c@{}}\includegraphics[scale=0.210]{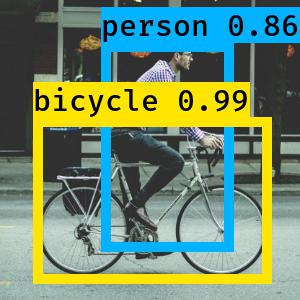}\\ \;\end{tabular}} & \multicolumn{1}{c|}{\begin{tabular}[c]{@{}c@{}}\includegraphics[scale=0.210]{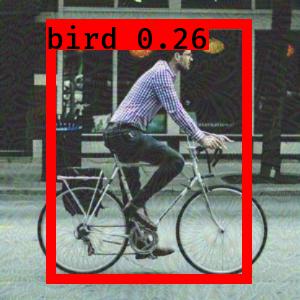}\\ \textbf{Transfer Succeeds}\end{tabular}} & \multicolumn{1}{c|}{\begin{tabular}[c]{@{}c@{}}\includegraphics[scale=0.210]{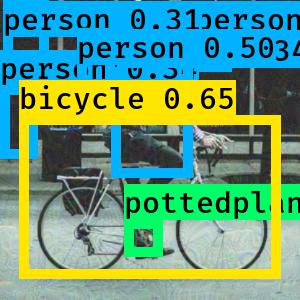}\\ \textbf{Transfer Succeeds}\end{tabular}} & \multicolumn{1}{c|}{\begin{tabular}[c]{@{}c@{}}\includegraphics[scale=0.210]{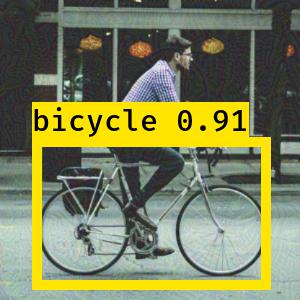}\\ \textbf{Transfer Succeeds}\end{tabular}} & \multicolumn{1}{c|}{\begin{tabular}[c]{@{}c@{}}\includegraphics[scale=0.210]{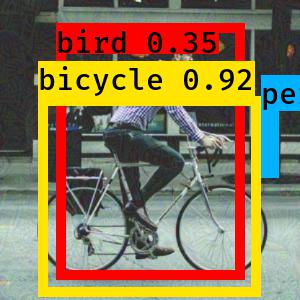}\\ \textbf{Transfer Succeeds}\end{tabular}} & \multicolumn{1}{c|}{\begin{tabular}[c]{@{}c@{}}\includegraphics[scale=0.210]{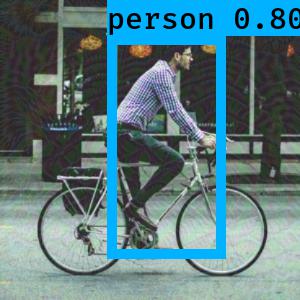}\\ \textbf{Transfer Succeeds}\end{tabular}} \\ \cline{3-9} 
		\end{tabular}
		\vspace{1em}
		\caption{Qualitative analysis (1st row) and adversarial transferability of \scheme{} attacks (2nd-4th rows).}
		\label{tab:online-attack-showcase}
	\end{table*}
	
	\subsection{Quantitative Analysis}
	Table~\ref{tab:overall_map} compares the mAP of each dataset and victim detector given benign examples (the $3$rd column) and adversarial examples (the $4$th-$8$th columns). 
	The four attacks are \scheme{}-vanishing, \scheme{}-fabrication, \scheme{}-mislabeling with ML and LL targets, and \scheme{}-universal. Compared to the benign mAP, all four attacks drastically reduce the mAP of the victim detector. For instance, \scheme{}-vanishing attacks on VOC and COCO break down the detection capability of the three YOLOv3 detectors: YOLOv3-D (VOC), YOLOv3-M (VOC) and YOLOv3-D (COCO), by reducing their mAP from $83.43\%$, $72.51\%$ and $54.89\%$ to $0.31\%$, $0.58\%$, and $0.51\%$ respectively. Also, the \scheme{}-mislabeling (LL) attacks collapse the mAP of all cases to less than $2\%$. Due to the space limit, we only report the comparison of the four victim detectors with respect to four \scheme{} attacks on the VOC dataset and the YOLOv3 detector with a Darknet53 backbone (YOLOv3-D) on the COCO dataset.
	
	It is worth noting that the above adversarial vulnerabilities are not limited to the detection capability for just a few classes but equally detrimental to any class supported by the victim detector. Figure~\ref{fig:exp-voc-yolov3-m} shows the average precision (AP) of all VOC classes on YOLOv3-M. Compared to the case with no attack (the benign case with blue curves), all four \scheme{} attacks are shockingly successful in bringing down the APs of the victim detector, and the \scheme{}-vanishing, \scheme{}-fabrication, \scheme{}-mislabeling attacks can drastically reduce the victim detector to very small or close to zero APs.
	
	From the ``\scheme{}-universal" column in Table~\ref{tab:overall_map} and Figure~\ref{fig:exp-voc-yolov3-m}(d), we make two additional observations. First, both show that our universal attacks reduce the mAP of all four detectors significantly with the most noticeable reduction in YOLOv3-D on VOC, with only a low mAP of $14.16\%$, compared with the high mAP of $83.43\%$ with no attack (benign mAP). Second, it also shows that the \scheme{}-universal attacks are less effective compared to the other three \scheme{} attacks. This is because the other three \scheme{} attacks are generated with per-input optimization, and the \scheme{}-universal attack generates a single universal perturbation through offline training for each victim detector, and then it is applied in real-time to any input sent to the victim detector without per-input fine-tuning optimization. For example, the \scheme{}-universal for VOC on YOLOv3-M generates the universal perturbation offline in $8$ hours but can be applied as a black-box attack to the victim detector in only $0.00136$ seconds, compared to $0.37$ seconds for \scheme{}-vanishing attack online. 
	
	\subsection{Qualitative Analysis}
	Given that all four attacks in \scheme{} significantly reduce mAP, we dedicate this subsection to perform qualitative analysis on the intrinsic behavior of each attack and explain how the detection capability of a victim detector is stripped off.
	
	The top part of Table~\ref{tab:online-attack-showcase} shows a test image (left) of a person riding a bicycle with the detection results made by SSD300 on benign (the ``Benign (No Attack)" column) and adversarial examples generated by four attacks (from ``\scheme{}-vanishing" to ``\scheme{}-universal" columns). Comparing the detection results on the benign example with the five adversarial counterparts (first row) for SSD300, we made a number of interesting observations. (1) The adversarial examples generated by \scheme{}-vanishing and \scheme{}-universal attacks both successfully remove the victim detector's capability in detecting objects (i.e., the person and the bicycle cannot be detected anymore), even though the \scheme{}-vanishing attack generates its adversarial perturbation tailored for this specific input image, while \scheme{}-universal uses a pretrained universal perturbation. (2) For the \scheme{}-fabrication attack, it fools the victim detector to give a large number of imprecise object detections (bounding boxes), all with high confidence, successfully tricks the victim detector. From the information retrieval perspective, our \scheme{}-vanishing and \scheme{}-universal attacks have a significant impact on the recall of the victim detector (unable to detect any object). In contrast, the \scheme{}-fabrication attack fools the victim detector to have a much lower precision because the detection results contain a larger number of bounding boxes without objects (``false objects" are everywhere). (3) The \scheme{}-mislabeling attacks (ML and LL) aim to disguise its true intent by making the victim detector to detect the same set of bounding boxes as those on benign examples under no attack (camouflage) but mislabel some or all detected objects with incorrect classes as demonstrated in both the ``\scheme{}-mislabeling (ML)" and the ``\scheme{}-mislabeling (LL)" columns in Table~\ref{tab:online-attack-showcase}. For instance, the person on a bicycle is mislabeled as a dog on a horse with high confidence under the \scheme{}-mislabeling (ML) attack. For the \scheme{}-mislabeling (LL) attack, the two objects are both mislabeled as a bus with at least $95\%$ confidence. Although camouflaged with bounding boxes, \scheme{}-mislabeling attacks successfully bring down the precision of the victim detector, because the detected bounding boxes are associated with wrong labels. 
	
	\subsection{\scheme{} Attack Transferability}\label{sec:transferability}We dedicate this subsection to study the transferability of \scheme{} attacks by generating adversarial examples on SSD300 and sending them to the other three detectors: SSD512, YOLOv3-D, and YOLOv3-M. We study whether the malicious perturbation generated from the attack to one victim detector can be effectively used as the black-box attack to fool the others. Table~\ref{tab:online-attack-showcase} (the $2$nd-$4$th rows) visualizes the detection results transferring the adversarial examples attacking SSD300 to the other three victim detectors. 
	
	First, with no attack, all three detectors can correctly identify the person and the bicycle upon receiving the benign input (the $1$st column). Second, the \scheme{} attacks have different degrees of adversarial transferability for different victim detectors under different attacks. Consider the victim detector SSD512, both \scheme{}-vanishing and \scheme{}-universal can perfectly transfer the attack to SSD512 with the same effect (i.e., no object is detected). For \scheme{}-fabrication, we observe that while the number of false objects is not as much as in the SSD300 case, a fairly large number of fake objects are wrongly detected by SSD512. The \scheme{}-mislabeling (LL) attack is transferred to SSD512 but with the object-fabrication effect instead, while the \scheme{}-mislabeling (ML) attack failed to transfer for this example. Now consider YOLOv3-D and YOLOv3-M, the \scheme{}-universal and the \scheme{}-mislabeling (LL) attacks are successful in transferability for both victim detectors but with different attack effects, such as wrong or additional bounding boxes or wrong labels. Also, the attacks from SSD300 can successfully transfer to YOLOv3-M with different attack effects compared to the attack results in SSD300. However, only the universal attack from SSD300 succeeds in transferring, but the other three \scheme{} attacks failed to transfer to YOLOv3-D for this example. Note that with adversarial transferability, the attacks are black-box, generated, and launched without any prior knowledge of the three victim detectors. 
	
	\section{Conclusion}\label{sec:conclusion}
	We have presented \scheme{}, a family of targeted adversarial objectness gradient attacks on deep object detection networks executing in edge systems. \scheme{} attacks enable adversaries to generate human-imperceptible perturbations, either by employing adversarial perturbation optimized for each input or by offline training a universal perturbation that is effective on any inputs to the victim detector. We also studied the adversarial transferability from one victim detector to others through black-box attacks. Through experiments on two benchmark datasets and four popular deep object detectors, we show the serious adversarial vulnerabilities of the representative deep object detection networks when deploying in edge systems.
	
\begin{figure}
		\centering
		\begin{subfigure}[t]{0.48\columnwidth}
			\centering
			\includegraphics[height=2.5cm, width=0.95\textwidth]{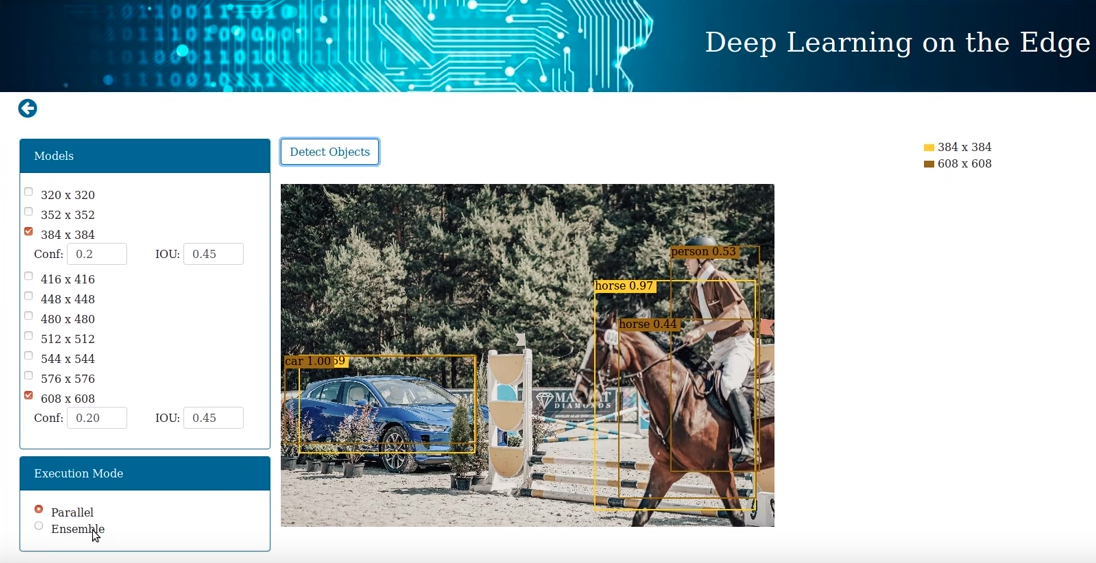}
			\caption{ A screenshot of the robust object detection system.}
		\end{subfigure}
		\hfill 
		\begin{subfigure}[t]{0.48\columnwidth}
			\centering
			\includegraphics[width=0.95\textwidth]{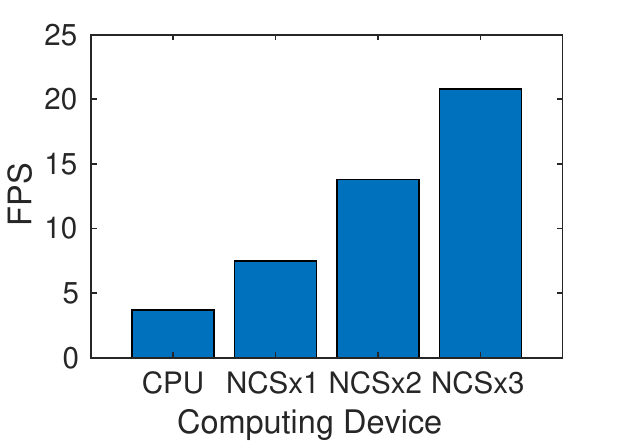}
			\caption{ The frame per second (FPS) running ensemble detections.}
		\end{subfigure}
		\vspace{-0.5em}
		\caption{ Robust real-time object detection using Intel NCS2.}
		\label{fig:ncs2_fps}
	\end{figure}

	From our experiences and experimental study on different victim detectors and on the adversarial transferability, we observe the divergence of attack effects on different detectors.  In general, an adversarial example attacking a victim model may not have the same adverse effect when used as a black-box attack based on its transferability. This is because the weak spot of attack transferability may vary from one detector to another trained by using a diverse DNN structure or diverse DNN algorithms as identified in~\cite{chow2019denoising,liu2019deep}. Our ongoing work is to develop diversity-enhanced ensemble object detection systems that promote strong robustness guarantees for defensibility and resilience against \scheme{} attacks. Our preliminary robustness study with Intel Neural Compute Stick 2 (NCS2) as the AI acceleration module on edge systems shows some encouraging results. Figure~\ref{fig:ncs2_fps} shows a robust object detection edge system developed at DiSL, Georgia Institute of Technology,  which offers real-time performance in an edge system using an ensemble of multiple object detectors.
	The alpha release of the open-source software package is accessible on GitHub at \url{https://github.com/git-disl/DLEdge}.
	
	\begin{acks}
		This research is partially sponsored by NSF CISE SaTC 1564097 and an IBM faculty award. We also thank Gage Bosgieter and the Intel Artificial Intelligence Developer Program for providing this research with the Intel Neural Compute Sticks. Any opinions, findings, and conclusions or recommendations expressed in this material are those of the author(s) and do not necessarily reflect the views of the National Science Foundation or other funding agencies and companies mentioned above.
	\end{acks}
	
	\bibliographystyle{acmref}
	\bibliography{ref}

\end{document}